\documentclass{article}

\usepackage{microtype}
\usepackage{graphicx}
\usepackage{subcaption}
\usepackage{booktabs} 

\usepackage{hyperref}

\usepackage[accepted]{icml2026}

\usepackage{amsmath}
\usepackage{amssymb}
\usepackage{mathtools}
\usepackage{amsthm}

\usepackage{hyperref}       
\usepackage{url}            
\usepackage{amsfonts}       
\usepackage{nicefrac}       
\usepackage{xcolor}         

\usepackage{graphicx}
\usepackage{booktabs}
\usepackage{multirow}
\usepackage{algorithm}
\usepackage{caption}
\usepackage{tcolorbox}
\usepackage{enumitem}

\usepackage{float}
\newtcolorbox{skillbox}[1]{
  colback=gray!4,
  colframe=gray!60,
  title=\textbf{#1},
  fonttitle=\small,
  fontupper=\small,
  boxrule=0.5pt,
  arc=2pt,
  left=5pt,
  right=5pt,
  top=5pt,
  bottom=5pt
}

\usepackage{xcolor}
\usepackage{subcaption}
\usepackage{wrapfig}
\newcommand{\PBtag}{\textcolor{teal!70!black}{\textbf{[PB]}}}
\newcommand{\GBtag}{\textcolor{red!70!black}{\textbf{[GB]}}}
\newcommand{\Dtag}{\textcolor{violet}{\textbf{[$d_i$]}}}

\usepackage{xcolor}
\newcommand{\Stag}{\textcolor{blue!75!black}{\textbf{[$s_i$]}}}
\newcommand{\Vtag}{\textcolor{orange!85!black}{\textbf{[prev\_v]}}}
\newcommand{\PGBtag}{\textcolor{magenta!75!black}{\textbf{[PGB]}}}

\usepackage[capitalize,noabbrev]{cleveref}

\theoremstyle{plain}

\theoremstyle{definition}

\theoremstyle{remark}

\usepackage[textsize=tiny]{todonotes}

\icmltitlerunning{AgentPSO: Evolving Agent Reasoning Skill via Multi-agent Particle Swarm Optimization}

\begin{document}

\twocolumn[
  \icmltitle{AgentPSO: Evolving Agent Reasoning Skill  via \\ Multi-agent Particle Swarm Optimization}



  \icmlsetsymbol{equal}{*}

  \begin{icmlauthorlist}
    \icmlauthor{Hyunmin Hwang}{equal,yyy}
    \icmlauthor{Jaemin Kim}{equal,yyy}
    \icmlauthor{Choonghan Kim}{yyy}
    \icmlauthor{Hangeol Chang}{yyy}
    \icmlauthor{Jong Chul Ye}{yyy}
  \end{icmlauthorlist}

  \icmlaffiliation{yyy}{Graduate School of Artificial Intelligence, Korea Advanced Institute of Science and Technology, Seoul, South Korea}

  \icmlcorrespondingauthor{Jong Chul Ye}{jong.ye@kaist.ac.kr}

  \icmlkeywords{Machine Learning, ICML}

  \vskip 0.3in
]



\printAffiliationsAndNotice{}  

\begin{abstract}
Multi-agent reasoning has shown promise for improving the problem-solving ability of large language models by allowing multiple agents to explore diverse reasoning paths.
However, most existing multi-agent methods rely on inference-time debate or aggregation, which can be vulnerable to incorrect peer influence and biased consensus.
Moreover, the agents themselves remain static, as their underlying reasoning skills do not evolve across tasks.
In this paper, we introduce \textbf{AgentPSO}, a particle-swarm-inspired framework for evolving multi-agent reasoning skills.
AgentPSO treats each agent as a particle-like reasoner whose state is a natural-language skill and whose velocity is a semantic update direction,
iteratively guiding agents toward higher-performing skill configurations.
Across training iterations, each agent updates its skill by combining its previous velocity, personal-best skill, global-best skill, and a self-reflective direction derived from peer reasoning trajectories. 
This enables agents to learn reusable reasoning behaviors by drawing on their own experience and on the strongest skills found by the population, without updating the parameters of the backbone language model. 
Experiments on mathematical and general reasoning benchmarks show that AgentPSO improves over static single-agent skills and test-time-only multi-agent reasoning baselines.
The evolved skills further transfer across benchmarks and to another backbone model, suggesting that AgentPSO captures reusable reasoning procedures rather than merely optimizing benchmark-specific prompts.
Code is publicly available at \url{https://github.com/HYUNMIN-HWANG/AgentPSO/}.
\end{abstract}

\section{Introduction}

``\textit{Groups are remarkably intelligent, and are often smarter than the smartest people in them.}'' \\
--- James Surowiecki, \textit{The Wisdom of Crowds}~\cite{surowiecki2004wisdom}

Large language models (LLMs) have shown remarkable progress in solving complex reasoning problems~\cite{openai2026gpt55,achiam2023gpt4,anthropic2026claude_models_overview}, 
yet their performance remains highly sensitive to how they are prompted and what reasoning skills they are instructed to use~\cite{wei2022chain,stepback,yao2023tot,selfconsistency,selfrefine,zhou2023leasttomost}.
As manually identifying optimal prompts and agent skills is labor-intensive, recent work has therefore explored automatic refinement of prompts, policies, and reasoning behaviors~\cite{grips,ape,opro,dspy,autodspy,gepa,memento_skills}. However, most prompt optimization and self-evolution methods focus on improving a single agent, prompt, or pipeline, which can limit exploratory diversity and lead to local optima~\cite{huang2023selfcorrectionyet}.

This motivates the use of multi-agent systems, where multiple agents provide diverse reasoning perspectives beyond the limitations of a single agent's self-contained evolution.
By critiquing one another and exposing alternative solution paths, multi-agent systems enable broader exploration of the reasoning space~\cite{multiagent_debate,mad_confidence,diverse_mad,groupdebate}.
However, existing multi-agent paradigms largely rely on inference-time collaboration, such as multi-round debate, which can be vulnerable to incorrect peer influence, biased consensus, and substantial computational overhead~\cite{kaesberg-etal-2025-voting-or-consensus,cui2026freemad}.

Crucially, while these methods improve the outcomes of the current discussion, they do not evolve the agents themselves or enable them to internalize reusable reasoning skills from prior interactions. Although memory-based approaches store and retrieve past lessons~\cite{lessons_learned,ling2025memad}, they mainly reuse externalized experiences at inference time instead of updating the agents' underlying reasoning skills. As a result, their ability to transfer general reasoning behaviors across benchmarks remains limited.

We therefore ask: \emph{Can multi-agent systems iteratively evolve their underlying reasoning skills through collective interaction, rather than merely refining the output for a single instance?}
We propose AgentPSO, a population-based framework for optimizing agent skills, inspired by the high-level structure of Particle Swarm Optimization (PSO)~\cite{PSO}.
In our framework, each agent is treated as a particle,
and its velocity~$v$ captures the direction of skill refinement.
At each iteration, agents update their skills by leveraging both their personal-best skill~$p$ and the global-best skill~$g$.
Through this population-guided update, AgentPSO exploits strong skills discovered by the swarm while maintaining diverse, agent-specific improvement trajectories.

However, directly applying the standard PSO rule to agent evolution is insufficient, as agents may move toward high-performing skill states without acquiring the reasoning principles that make them effective~\cite{freitas2020particle-review,gad2022particle-review,zhang2019CLPSO-LOT}. 
AgentPSO therefore introduces a self-reflective direction, which allows each agent to analyze peer reasoning trajectories and extract reusable lessons instead of simply copying stronger skills. 
This reflection mechanism allows agents to internalize procedural reasoning patterns that generalize beyond individual task instances.
Over iterations, the initially diverse agents evolve into a collectively stronger population.
At test time, the evolved agents solve problems independently and aggregate their answers through majority voting, achieving strong performance without costly multi-round interaction.

We evaluate AgentPSO on mathematical and general reasoning benchmarks.
Our experiments show that AgentPSO outperforms single-agent prompting methods and debate-based multi-agent baselines. 
Further analyses show that the evolved skills transfer across benchmarks, improve progressively over iterations, and benefit substantially from the self-reflective direction.

Our contributions are summarized as follows:
\begin{itemize}
    \item We introduce \textbf{AgentPSO}, a PSO-inspired framework that evolves multi-agent reasoning skills by updating agents through personal-best, global-best, and self-reflective directions.
    \item We demonstrate that AgentPSO yields persistent evolution of agent capabilities, achieving stronger performance than single-agent and debate-based multi-agent baselines without requiring costly test-time debate.
    \item We demonstrate that evolved skills transfer across benchmarks, indicating that AgentPSO learns general reasoning behaviors rather than merely memorizing task-specific knowledge.
\end{itemize}

\begin{figure*}[t]
  \centering
  \includegraphics[
    width=1\textwidth,
    trim=1cm 5cm 0 0,
    clip
  ]{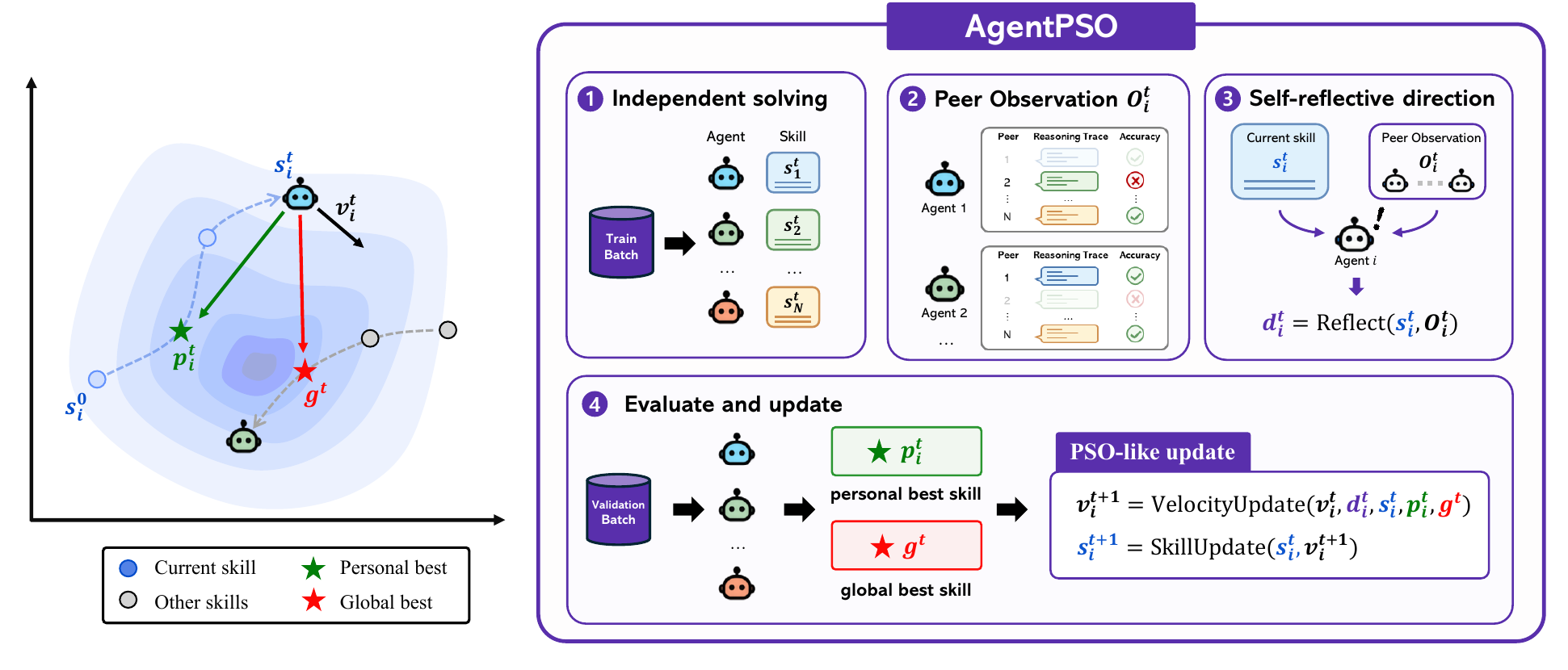}  
\caption{
\textbf{Overview of AgentPSO.}
Each agent independently solves the training batch with its current skill, producing answers and reasoning traces.
Peer observations summarize other agents' reasoning traces and correctness, from which each agent derives a self-reflective direction for skill improvement. 
This direction is combined with guidance from the personal-best and global-best skills through a PSO-like update.
}\label{fig:overall_framework}
\vspace{-0.5cm}
\end{figure*}

\section{Related Works}
\label{sec:related_works}

\paragraph{Multi-Agent Collaboration Approaches.}

Individual language models are often limited in their ability to explore diverse reasoning trajectories~\cite{huang2023selfcorrectionyet}, motivating multi-agent collaboration in which agents exchange information and refine their solutions collectively. 
Prior work has shown that multi-agent debate can improve reasoning by enabling agents to critique one another and expand the space of possible solutions~\cite{multiagent_debate,diverse_mad}. 
Later studies improve reliability through confidence-aware interaction~\cite{mad_confidence}, multidimensional assessment~\cite{mmad}, cross-agent reward signals~\cite{comas}, and tournament-style comparison~\cite{catarena}. 
Other approaches augment collaboration with external memory, allowing agents to retrieve lessons from prior debates for future tasks~\cite{groupdebate,lessons_learned}. 
However, these methods remain vulnerable to negative peer influence~\cite{kaesberg-etal-2025-voting-or-consensus,cui2026freemad,weng2025conformityllm}, require costly inference-time collaboration, and do not directly improve agents' underlying capabilities across tasks.
By contrast, we formulate multi-agent collaboration as agent skill evolution, where agents improve by learning from each other’s reasoning trajectories.

\paragraph{Self-Evolution in Language Models.}
Recent research has increasingly aimed to make language models self-evolving by iteratively adapting their behavior to better fit a target task~\cite{surveyselfevolvingagentspath}. 
Early efforts focus on prompting, where task-relevant reasoning depends heavily on the prompts provided to the model~\cite{wei2022chain,selfrefine,decomposed,shinn2023reflexion}. 
Subsequent works therefore explore prompt optimization to automatically discover more effective prompts~\cite{grips,ape,opro,dspy,autodspy,promptbreeder,gepa}. 
More recent studies further extend this line of research from prompt optimization to the iterative refinement of reasoning structures, workflows, agent architectures, and externalized skills through feedback and self-improvement~\cite{self_discover,debflow,adas,dgm,comas,medagentsim,wei2025evomemory,wang2025AWM,memento_skills}.
Our method follows this direction, but focuses on population-level multi-agent skill evolution rather than optimizing a single component, such as a prompt, agent, or pipeline.
Specifically, each agent updates its own skill based on its current state, a self-reflective update direction, its personal-best skill, and the global-best skill discovered across the population.

\paragraph{Particle Swarm Optimization.}
We interpret multi-agent skill evolution through the lens of Particle Swarm Optimization (PSO)~\cite{PSO}, in which each particle is updated based on its current state, its personal best, and the swarm’s global best. 
PSO has been widely applied to diverse optimization problems, including in language model optimization~\cite{llmPSO,gad2022particle-review}. 
We extend this perspective to multi-agent evolution by treating each agent as a particle-like reasoner with an evolving skill state. 
Each agent is thus guided by both its own personal-best skill and the strongest skill discovered across the population, enabling improvement at both the individual and collective levels. 
SwarmAgentic~\cite{zhang2025swarmagentic} also applies swarm intelligence to automated agentic system generation by jointly optimizing agent functionality and collaboration structures.
However, standard PSO alone can be limited on complex search problems, where effective exploration remains difficult~\cite{freitas2020particle-review,zhang2019CLPSO-LOT}. 
To address this limitation, our method extends the PSO-style update with a self-reflective {\em semantic} direction term by leveraging the capabilities of LLMs, encouraging broader exploration while steering agents toward better skill configurations.

\section{Problem Setup}
\label{sec:problem_setup}

We consider a population of $N$-LLM agents $A=\{A_1,A_2,\ldots,A_N\}$ that jointly evolve their reasoning skills over problem-solving iterations. 
At iteration $t$, each agent $A_i$ has a current skill state $s_i^t \in \mathcal{S}$, where $\mathcal{S}$ denotes the space of valid skill states.
A skill state specifies how the agent approaches reasoning problems; different skill states lead to different reasoning trajectories.
The agents are initialized with diverse skill instructions, 
allowing the population to start from multiple reasoning strategies rather than a single shared prompt.

We define $F(s)$ as the validation accuracy obtained when an agent solves a validation batch using skill $s$.
Our objective is to obtain a population of high-performing skill states $\{s_i^\ast\}_{i=1}^N$ that improve both individual agent performance and collective prediction.
We optimize only the skill instructions, without updating the internal parameters of the backbone LLM.
Because the optimized object is a reusable instruction rather than a task-specific answer, 
the resulting skills are expected to improve performance on unseen problems. 
During optimization, training batches are used to generate peer observations and update directions, whereas validation batches are used to select personal-best and global-best skills.

Our goal is not to optimize each skill independently.
If each agent only updates its own instruction, the search remains limited to that agent's local trajectory.
Instead, we use the population as a source of comparative evidence.
When multiple agents solve the same batch, their reasoning traces reveal which strategies succeed, which fail, and which reasoning behaviors are useful across problems.
Thus, skill evolution is driven not only by an agent's own experience, but also by behavioral evidence extracted from peer trajectories.

To perform this population-based skill search, we adopt a swarm-inspired formulation.
Each agent maintains its current skill, its own search history, the personal-best skill it has found so far, and the global-best skill found by the population.
In addition, each agent uses peer reasoning trajectories to infer how its current skill should be revised.
We treat each agent $A_i$ as a particle whose position is its current skill state $s_i^t$.
Unlike standard particle-based optimization, however, this position is not a numerical vector but a natural-language instruction.

\section{AgentPSO}
\label{sec:method}

\begin{table}[t]
\centering
\caption{
\textbf{Conceptual mapping between standard PSO and AgentPSO.}
AgentPSO maps high-level PSO concepts to natural-language skill optimization, replacing numerical vector arithmetic with LLM-based semantic update operators.
}
\label{tab:pso_agentpso}
\resizebox{\columnwidth}{!}{ 
\renewcommand{\arraystretch}{1.08}
\begin{tabular}{lcc}
\toprule
 & \textbf{Standard PSO} & \textbf{AgentPSO} \\
\midrule
State & Continuous position $x_i^t$ & Skill state $s_i^t$ \\
Direction & Vector velocity $v_i^t$ & Semantic velocity $v_i^t$ \\
Velocity update 
& $\omega v_i^t + c_1 r_1(p_i^t-x_i^t) + c_2 r_2(g^t-x_i^t)$ 
& $\mathrm{VelocityUpdate}(v_i^t,d_i^t,s_i^t,p_i^t,g^t)$ \\
State update 
& $x_i^t + v_i^{t+1}$ 
& $\mathrm{SkillUpdate}(s_i^t,v_i^{t+1})$ \\
\bottomrule
\end{tabular}
}
\vspace{-0.6cm}
\end{table}

AgentPSO is a PSO-inspired framework for multi-agent reasoning skill evolution.
Rather than directly instantiating numerical PSO updates, AgentPSO reformulates the personal-best and population-best mechanisms as semantic skill-revision operators, enabling natural-language skill optimization.
AgentPSO adapts this structure to natural-language skill optimization by treating each skill instruction as a particle state and each textual revision direction as a semantic velocity. Table~\ref{tab:pso_agentpso} summarizes this correspondence.

Each training iteration follows four steps:
(i) All agents solve the same training batch independently using their current skills.
(ii) Each agent observes the answer correctness and the reasoning traces produced by the population and derives a self-reflective update direction.
(iii) Each agent combines this direction with its previous semantic velocity, its personal-best skill, and the global-best skill to produce a new semantic velocity, which is then used to rewrite its current skill.
(iv) The updated skills are evaluated on a validation batch, and the validation scores are used to update the personal-best and global-best skills. We visualize the overall pipeline of AgentPSO in Figure~\ref{fig:overall_framework}.

\subsection{Independent Solving and Peer Observation}
\label{subsec:peer_observation}

At iteration $t$, all agents first solve the same training batch independently using their current skills $s_i^t$
{\em without} inter-agent communication.
This independent solving phase allows us to evaluate the reasoning behavior induced by each skill before it is affected by peer interaction.
In contrast, debate-based methods allow agents to exchange arguments during inference, which entangles their outputs and makes it harder to attribute the final result to a particular skill.
In addition, AgentPSO maintains this non-interactive design at test time. 
The evolved agents solve problems in parallel and aggregate their answers by majority voting, avoiding the multi-round communication overhead of debate-based inference.

After all agents complete the batch,
we construct a peer observation $O_i^t$ for each agent.
This observation contains the agent's own output and the peer agents' outputs, including final answers, reasoning traces, and correctness labels, while excluding the peer agents' skill instructions.
This design encourages agents to learn from peer reasoning behaviors rather than copying peer skill texts.
Consequently, each agent can identify what made another trajectory successful (\textit{e.g.}, checking intermediate assumptions or decomposing the problem more carefully), and use these observations to improve its own skill.

\setlength{\textfloatsep}{10pt}
\begin{algorithm}[t]
\caption{AgentPSO}
\label{alg:agentpso}
    \begin{algorithmic}[1]
    \REQUIRE Validation accuracy $F$, initial skills $\{s_i^0\}_{i=1}^N$, iterations $T$, margin $\epsilon$
    \STATE $\{v_i^0, p_i^0\} \gets \{\text{``''}, s_i^0\}$ for all $i$
    \STATE $g^0 \gets s_{j^\star}^0$, where $j^\star=\arg\max_i F(s_i^0)$
    \FOR{$t=0$ to $T-1$}
      \STATE Each agent solves the training batch using $s_i^t$
      \FOR{$i=1$ to $N$}
          \STATE Construct peer observation $O_i^t$
          \STATE $d_i^t \gets \mathrm{Reflect}(s_i^t, O_i^t)$
          \STATE $v_i^{t+1} \gets \mathrm{VelocityUpdate}(v_i^t, d_i^t, s_i^t, p_i^t, g^t)$
          \STATE $s_i^{t+1} \gets \mathrm{SkillUpdate}(s_i^t, v_i^{t+1})$
      \ENDFOR
      \STATE Evaluate $\{s_i^{t+1}\}$ on a validation batch
      \STATE Update each $p_i^{t+1}$ with $s_i^{t+1}$ only if its validation score improves by more than $\epsilon$
      \STATE Update $g^{t+1}$ only when it improves global-best validation score by more than $\epsilon$
    \ENDFOR
    \STATE \textbf{return} $\{p_i^T\}_{i=1}^N$ and $g^T$
    \end{algorithmic}
\end{algorithm}

\subsection{PSO-Inspired Skill Evolution}
\label{subsec:skill_update}

Having constructed the peer observations, the next step is to update each agent's skill by integrating its own experience with the population's discoveries. 
To structure this integration, AgentPSO follows the update process based on the mechanics of Particle Swarm Optimization (PSO). 

Standard PSO maintains, for each particle, its current position $x_i^t$, velocity $v_i^t$, personal-best position $p_i^t$, and swarm-level global-best position $g^t$.
It updates each particle in two stages: first, it updates the velocity using the previous velocity, the particle's personal-best position, and the swarm's global-best position; then, it updates the particle's position by moving it according to the new velocity.

However, applying this update structure to natural-language skills requires an additional source of direction.
In continuous PSO, the {vector} differences $p_i^t-x_i^t$ and $g^t-x_i^t$ provide explicit directions toward better positions.
In natural-language skill space, however, knowing that a personal-best or global-best skill performed well does not directly explain which reasoning behavior should be added, removed, or strengthened.
Without such behavioral guidance, the update tends to imitate the best skill state rather than learning a reusable reasoning principle.

To address this, AgentPSO introduces a self-reflective direction $d_i^t$ derived from the peer observation $O_i^t$:
\begin{equation}
d_i^t = \mathrm{Reflect}(s_i^t, O_i^t).
\label{eq:reflect}
\end{equation}
This direction tells the agent how to improve its skill, given how its own behavior compares with peers' on the same batch.
Thus, $d_i^t$ provides a concrete behavioral update direction without directly copying another agent's skill instruction.

AgentPSO retains the PSO two-stage update within the natural-language skill space. For agent $A_i$, the current skill $s_i^t$ functions as the particle position, representing the candidate skill being optimized. 
The semantic velocity $v_i^t$ acts as the particle velocity, representing the textual revision direction applied at the next step. 
The personal-best skill $p_i^t$ and the global-best skill $g^t$ correspond to the best skill previously found by $A_i$ and the best skill discovered by the population, respectively.

The updated semantic velocity is generated as
\begin{equation}
\label{eq:velocity}
v_i^{t+1}
=
\mathrm{VelocityUpdate}\bigl(
v_i^t, d_i^t, s_i^t, p_i^t, g^t
\bigr).
\end{equation}
Here, $\mathrm{VelocityUpdate}$ replaces numerical vector differences with a semantic update operator.
Given the previous velocity, the self-reflective direction, the current skill, and the personal-best and global-best skill references, the LLM synthesizes a new textual velocity that describes how the current skill should be revised.

Finally, AgentPSO performs the analogue of the PSO position update by rewriting the current skill according to the updated velocity:
\begin{equation}
\label{eq:skill}
s_i^{t+1}
=
\mathrm{SkillUpdate}\bigl(
s_i^t, v_i^{t+1}
\bigr).
\end{equation}

Rather than adding a numerical velocity vector to a numerical position vector, $\mathrm{SkillUpdate}$ semantically rewrites the current skill.
The updated skill builds on the agent's current skill rather than starting from scratch, and remains applicable to new problems.
{Detailed prompt templates for $\mathrm{Reflect}$, $\mathrm{VelocityUpdate}$, and $\mathrm{SkillUpdate}$ are provided in Appendix~\ref{app:prompts}.}

\subsection{Validation-Based Best-State Selection}
\label{subsec:best_state_update}

After generating the updated skills, AgentPSO evaluates each $s_i^{t+1}$ on a validation batch using the validation accuracy $F$.
If $F(s_i^{t+1})$ exceeds the score of the current personal-best skill $p_i^t$ by more than $\epsilon$, we set $p_i^{t+1}=s_i^{t+1}$; otherwise, we keep $p_i^{t+1}=p_i^t$.
Likewise, the global-best skill is updated only if the strongest updated skill improves the current global-best score by more than $\epsilon$.
This margin $\epsilon$ prevents small or noisy fluctuations from immediately replacing previously effective skills.

At initialization, the semantic velocities are set to empty update directions, $v_i^0=\text{``''}$.
The personal-best skills are initialized as $p_i^0=s_i^0$, and the initial global-best skill $g^0$ is selected as the highest-scoring initial skill on the validation batch. Algorithm~\ref{alg:agentpso} summarizes the overall procedure.

\section{Experimental Setup}
\label{sec:experiments}

\paragraph{Datasets.}
We evaluate AgentPSO on five benchmarks covering mathematical and general reasoning.
(1) For mathematical reasoning, we use DeepMath~\cite{deepmath}, MATH~\cite{math500}, AIME25~\cite{aime25}, and Minerva~\cite{minerva}, which require multi-step reasoning, symbolic manipulation, and numerical problem solving. 
The resulting optimized agent skills are evaluated on the DeepMath test set and further applied to out-of-distribution mathematical benchmarks, including MATH, AIME25, and Minerva.
(2) For general reasoning, we use BigBenchHard (BBH)~\cite{bbh}, a challenging benchmark composed of 23 diverse reasoning tasks, including logical, symbolic, commonsense, and multi-step reasoning. 
We randomly split BBH into training, validation, and test sets and report the final performance on the BBH test split.
Further details on the datasets and splits are provided in Appendix~\ref{app:datasets}.

\begin{table*}[t]
\centering
\caption{
{
\textbf{Comparison with single-agent, multi-agent, and self-evolving baselines across GPT-5.4-mini and Qwen3-32B.}
All reported values denote accuracy. AgentPSO achieves the best average performance with GPT-5.4-mini and remains competitive with Qwen3-32B; the GPT--Qwen transfer variant obtains the highest average accuracy in the Qwen3-32B settings.
}
}
\label{tab:combined_main_qwen_results}
\scriptsize
\setlength{\tabcolsep}{2.5pt}
\renewcommand{\arraystretch}{1.12}
\resizebox{\textwidth}{!}{%
\begin{tabular}{ll|cccccc|cccccc}
\toprule
\multirow{2}{*}{\textbf{Category}} 
& \multirow{2}{*}{\textbf{Method}}
& \multicolumn{6}{c|}{\textbf{GPT-5.4-mini}}
& \multicolumn{6}{c}{\textbf{Qwen3-32B}} \\
\cmidrule(lr){3-8} \cmidrule(lr){9-14}
& 
& \textbf{DeepMath}
& \textbf{MATH}
& \textbf{AIME25}
& \textbf{Minerva}
& \textbf{BBH}
& \textbf{Avg.}
& \textbf{DeepMath}
& \textbf{MATH}
& \textbf{AIME25}
& \textbf{Minerva}
& \textbf{BBH}
& \textbf{Avg.} \\
\midrule

\multirow{2}{*}{\shortstack[c]{Vanilla \\ Single-Agent}}
& Chain-of-Thought 
& 68.00 & 90.40 & 40.00 & 38.60 & 85.50 & 64.50
& 60.00 & 77.20 & 10.00 & 37.50 & 71.50 & 51.24 \\
& Step-Back Prompting 
& 64.50 & 90.40 & 50.00 & 38.60 & 82.50 & 65.20
& 51.50 & 63.80 & 6.67 & 37.87 & 66.50 & 45.27 \\

\midrule
\multirow{4}{*}{\shortstack[c]{Advanced \\ Single-Agent}}
& Self-Refine      
& 70.50 & 91.60 & 53.33 & 34.56 & 82.50 & 66.50
& 62.00 & 78.40 & 6.67 & \textbf{41.54} & 71.00 & 51.92 \\
& Reflection       
& 69.50 & 90.80 & 40.00 & 38.24 & 83.00 & 64.31
& 57.00 & 74.00 & 13.33 & 41.18 & 74.00 & 51.90 \\
& Self-Consistency 
& 69.50 & 88.80 & \textbf{60.00} & 27.21 & 86.50 & 66.40
& 73.50 & 81.40 & 20.00 & 36.40 & \textbf{85.00} & 59.26 \\
& ToT       
& 63.00 & 81.60 & 50.00 & 31.99 & 78.50 & 61.02
& 55.50 & 68.00 & 16.67 & 25.00 & 78.00 & 48.63 \\

\midrule
\multirow{3}{*}{\shortstack[c]{Multi-Agent}}
& MAD       
& 69.00 & \textbf{96.00} & 56.67 & 32.72 & 84.00 & 67.68
& 71.00 & 86.00 & \textbf{26.67} & 40.44 & 81.50 & 61.12 \\
& MoA       
& 70.00 & 95.60 & 56.67 & 27.57 & 77.50 & 65.47
& 73.50 & 85.20 & 23.33 & 38.97 & 82.00 & 60.60 \\
& DMAD     
& 72.00 & 95.00 & 56.67 & 35.29 & 85.00 & 68.79
& 74.00 & 87.00 & \textbf{26.67} & 39.71 & 82.00 & 61.88 \\

\midrule
\multirow{3}{*}{\shortstack[c]{Self-Evolving}}
& DSPy     
& 76.00 & 92.40 & 53.33 & 35.66 & 84.50 & 68.38
& 71.50 & 84.40 & 16.67 & 36.40 & 79.00 & 57.59 \\
& GEPA    
& \textbf{79.50} & 91.40 & 43.33 & 41.91 & 86.50 & 68.53
& 72.00 & 86.00 & 23.33 & 38.97 & 77.50 & 59.56 \\
& ADAS     
& 61.00 & 79.40 & 53.33 & 30.51 & 83.00 & 61.45
& 53.50 & 65.00 & 23.33 & 28.68 & 75.50 & 49.20 \\

\midrule
\multirow{2}{*}{\shortstack[c]{AgentPSO}}
& AgentPSO 
& \textbf{79.50} & 94.60 & 56.67 & \textbf{42.65} & \textbf{87.00} & \textbf{72.08}
& 79.50 & 87.00 & 23.33 & 35.66 & 82.50 & 61.60 \\
& AgentPSO~(GPT--Qwen)
& -- & -- & -- & -- & -- & --
& \textbf{81.00} & \textbf{87.20} & \textbf{26.67} & 37.13 & 83.00 & \textbf{63.00} \\

\bottomrule
\end{tabular}
}
\end{table*}

\begin{figure*}
  \centering
  \includegraphics[
    width=1\linewidth,
    trim=0 11.2cm 0 0,
    clip
  ]{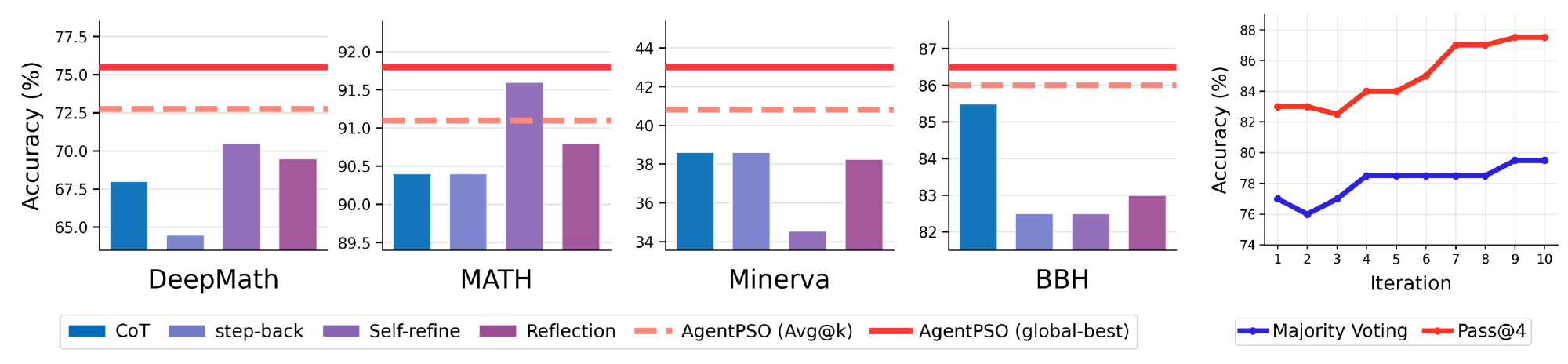}  
  
\caption{
\textbf{Progressive improvement of AgentPSO-evolved skills.}
(Left) Evolved personal-best and global-best skills outperform the initial skills on average.
(Right) DeepMath performance improves over iterations, showing gradual refinement of the agent population.
}
  \label{fig:evolved_agent_analysis}
  \vspace{-0.8em}
\end{figure*}

\paragraph{Baselines.}
We compare AgentPSO with baselines from four main categories. The implementation details of all baselines are described in Appendix~\ref{app:baseline_details}:
\begin{itemize}
    \item \textbf{Vanilla single-agent methods.}
    We include Chain-of-Thought Prompting (CoT)~\cite{wei2022chain} and Step-Back Prompting~\cite{stepback} as basic single-agent reasoning baselines.

    \item \textbf{Advanced single-agent methods.}
    We include Self-Refine~\cite{selfrefine}, Reflection~\cite{shinn2023reflexion}, Self-Consistency~\cite{selfconsistency}, and Tree-of-Thoughts~\cite{yao2023tot}, which improve single-agent reasoning through refinement, reflection, sampling, or structured search.

    \item \textbf{Multi-agent methods.}
    We compare against MAD~\cite{multiagent_debate}, MoA~\cite{MOA}, and DMAD~\cite{diverse_mad}, which leverage multiple agents to improve reasoning through debate, aggregation, or diversity.
    {
    \item \textbf{Self-evolving methods.}
    We further include DSPy~\cite{dspy} and GEPA~\cite{gepa} as prompt optimization baselines, and ADAS~\cite{adas} as an automated agentic-system design baseline.
    }
    
\end{itemize}

\paragraph{Implementation Details.}
We implement AgentPSO with ChatGPT (gpt-5.4-mini) as the backbone model. 
All experiments use four agents initialized with different CoT-based reasoning methods. 
We optimize agent skill states for {100 training samples over 10 iterations}, without updating backbone model parameters.
Detailed hyperparameter settings and prompts are provided in Appendices~\ref{app:pso_hyperparameters} and~\ref{app:prompts}.
At each validation step, we evaluate the current agent skills on 20 validation samples. 
To reduce overfitting to a single small validation subset, we rotate validation subsets during optimization, as described in Appendix~\ref{app:datasets}.

\section{Main Results}
\label{sec:main_results}

\subsection{AgentPSO Improves Multi-Agent Reasoning}
\label{subsec:results_evolve}

\begin{table*}[t]
\centering

\begin{minipage}[t]{0.4\linewidth}
\centering
\caption{\textbf{Cross-benchmark transfer of learned reasoning skills.}
}
\label{tab:cross_dataset_transfer}
\small
\setlength{\tabcolsep}{5pt}
\renewcommand{\arraystretch}{1.12}
\resizebox{\linewidth}{!}{%
\begin{tabular}{l|l|ccc}
\toprule
\textbf{Dataset} 
& \textbf{Method} 
& \textbf{Accuracy}
& \textbf{Pass@$k$} 
& \textbf{Avg@$k$} \\
\midrule
\multirow{3}{*}{DeepMath}
& MAD 
& 69.00 
& 70.50 
& 68.00 \\
& AgentPSO$_{\text{DeepMath}}$
& 79.50 
& 87.00 
& 72.75 \\
& AgentPSO$_{\text{BBH}}$
& 75.00 
& 81.00 
& 66.13 \\
\midrule
\multirow{3}{*}{BBH}
& MAD 
& 84.00 
& 87.00 
& 81.88 \\
& AgentPSO$_{\text{BBH}}$
& 87.00 
& 93.00 
& 85.88 \\
& AgentPSO$_{\text{DeepMath}}$
& 85.50 
& 90.50 
& 82.75 \\
\bottomrule
\end{tabular}
}
\end{minipage}
\hfill
\begin{minipage}[t]{0.59\linewidth}
\centering

\caption{
\textbf{Cross-model transfer of AgentPSO-learned skills from GPT to Claude.}
}
\label{tab:gpt_skill_to_claude}
\small
\setlength{\tabcolsep}{4pt}
\renewcommand{\arraystretch}{1.12}
\resizebox{\linewidth}{!}{%
\begin{tabular}{l|ccccc|c}
\toprule
\textbf{Method}
& \textbf{DeepMath}
& \textbf{MATH}
& \textbf{AIME25}
& \textbf{Minerva}
& \textbf{BBH}
& \textbf{Avg.} \\
\midrule
Chain-of-Thought & 60.00 & 90.60 & 30.00 & 36.03 & 47.50 & 52.83 \\
Step-Back Prompting & 63.50 & 89.80 & 36.67 & \textbf{36.40} & 51.00 & 55.47 \\
Self-Refine      & 61.00 & \textbf{91.60} & 30.00 & 34.56 & 50.00 & 53.43 \\
Reflection       & 60.00 & 90.20 & 36.67 & 35.66 & 55.00 & 55.51 \\
\midrule
\textbf{AgentPSO (GPT$\rightarrow$Claude)}
& \textbf{70.50} & 91.40 & \textbf{40.00} & 33.46 & \textbf{56.50} & \textbf{58.37} \\
\bottomrule
\end{tabular}
}
\end{minipage}
\vspace{-0.5em}
\end{table*}

\begin{table*}[t]
\centering

\begin{minipage}[t]{0.58\textwidth}
\centering
\captionof{table}{
\textbf{Reusing AgentPSO-trained skills in multi-agent inference methods.}
AgentPSO-trained skills are plugged into MAD and MoA at test time.}
\label{tab:agentpso_test_time_collaboration}
\scriptsize
\setlength{\tabcolsep}{5pt}
\renewcommand{\arraystretch}{0.9}
\resizebox{\linewidth}{!}{%
\begin{tabular}{l|cc|cc}
\toprule
\textbf{Dataset} 
& \textbf{MAD}
& \textbf{AgentPSO + MAD}
& \textbf{MoA}
& \textbf{AgentPSO + MoA} \\
\midrule
DeepMath     
& 69.00  & 76.50 (+7.50)  & 70.00  & 72.50 (+2.50) \\
Minerva      
& 32.72 & 36.40 (+3.68) & 27.57 & 29.78 (+2.21) \\
BigBenchHard 
& 84.00  & 84.00 (+0.00)  & 77.50  & 82.50 (+5.00)\\
\bottomrule
\end{tabular}
}
\end{minipage}
\hfill
\begin{minipage}[t]{0.4\textwidth}
\centering
\captionof{table}{
\textbf{Effect of the velocity term on skill stability.}
Semantic drift is computed between consecutive skill states using embedding-based cosine distance.
}
\label{tab:velocity_ablation}
\scriptsize
\setlength{\tabcolsep}{4pt}
\renewcommand{\arraystretch}{1.12}
\resizebox{\linewidth}{!}{%
\begin{tabular}{l|c|ccc|c}
\toprule
\textbf{Method}
& \textbf{Acc.(\%)}
& \textbf{Mean}
& \textbf{Median}
& \textbf{Max}
& \textbf{High(\%)} \\
\midrule
w/o velocity
& 78.0
& 0.0965
& 0.0562
& 0.4486
& 32.5 \\
w/ velocity
& \textbf{79.5}
& \textbf{0.0868}
& \textbf{0.0488}
& \textbf{0.4153}
& \textbf{15.0} \\
\bottomrule
\end{tabular}
}
\end{minipage}

\vspace{-0.6em}
\end{table*}

\paragraph{Main performance comparison.}

To evaluate the effectiveness of AgentPSO's skill evolution, we compare it against vanilla single-agent, advanced single-agent, multi-agent, and self-evolving baselines across GPT-5.4-mini and Qwen3-32B backbones (Table~\ref{tab:combined_main_qwen_results}).
For GPT, AgentPSO achieves the best average performance across the five benchmarks (72.08\%), outperforming single-agent prompting methods, static multi-agent baselines, and self-evolving methods.
This pattern suggests that AgentPSO does not merely improve the in-distribution optimization benchmark, but also learns skills that generalize to different mathematical and general reasoning benchmarks.

We further examine whether AgentPSO can optimize agent skills with Qwen3-32B as the backbone.
In the Qwen-only setting, all stages of AgentPSO, including reflection, velocity construction, and skill update, are performed using Qwen3-32B.
AgentPSO achieves an average accuracy of 61.60\%, 
which is competitive with Qwen-based multi-agent baselines and outperforms Qwen-based self-evolving methods.
This result shows that AgentPSO is not limited to proprietary models and remains effective with an open-weight backbone.
Furthermore, AgentPSO~(GPT--Qwen), where GPT performs the skill-evolution steps in Eq.~(\ref{eq:reflect})--(\ref{eq:skill}) while Qwen3-32B is used for inference, achieves the best Qwen average accuracy of 63.00\%.
This further suggests that AgentPSO benefits from a stronger update model, highlighting the importance of reflection and skill revision quality.

\paragraph{Progressive improvement of evolved agent skills.}
To examine whether AgentPSO genuinely improves agent skills through iterative evolution, we analyze the personal-best and global-best skills over the optimization process.
Figure~\ref{fig:evolved_agent_analysis} (Left) compares the initial agent skills with the learned personal-best and global-best skills.
After optimization, both the average personal-best agents and the global-best agent outperform the initial agents on average.
We find that the global-best agent achieves the strongest single-agent performance, indicating AgentPSO drives the agent population toward more effective skill configuration.
Figure~\ref{fig:evolved_agent_analysis} (Right) shows how performance changes across optimization iterations.
AgentPSO improves the population gradually rather than converging to a local optimum.
As the iterations progress, the collective performance of the saved personal-best agents steadily improves, with the final iterations achieving the strongest results.

\subsection{AgentPSO Learns Transferable Reasoning Skills}
\label{subsec:transfer_results}

\paragraph{Cross-domain transfer.}
To test whether AgentPSO discovers generalizable reasoning rather than overfitting to a particular dataset, 
Table~\ref{tab:cross_dataset_transfer} evaluates cross-benchmark transfer of AgentPSO-learned skills, where Pass@$k$ denotes whether at least one of the $k$ agents produces the correct answer and Avg@$k$ denotes the average accuracy across the $k$ individual agents.
We denote AgentPSO skills optimized on BigBenchHard and DeepMath as AgentPSO$_{\text{BBH}}$ and AgentPSO$_{\text{DeepMath}}$, respectively.
Although skills optimized on the target benchmark achieve the strongest performance, cross-trained AgentPSO skills still outperform the static MAD baseline on both DeepMath and BBH.
This trend is observed not only in majority-vote accuracy but also in pass@$k$, suggesting that the evolved agent population retains useful reasoning behaviors even when transferred to a different benchmark.
These results suggest that AgentPSO learns reasoning procedures that transfer 
across reasoning domains,
not just skills tied to the training data.

\paragraph{Cross-model transfer.}
Table~\ref{tab:gpt_skill_to_claude} 
evaluates whether AgentPSO's skills transfer across different backbone LLMs.
We obtain the final skill population optimized through AgentPSO using GPT, then we directly apply these skills to Claude~(claude-haiku-4-5-20251001) at test time without any additional training or adaptation. 
The single-agent baselines report the performance of Claude using the initial skill prompts, including Chain-of-Thought, Step-Back Prompting, Self-Refine, and Reflection.
AgentPSO~(GPT$\rightarrow$Claude) reports the performance of Claude when using the GPT-optimized AgentPSO skill.
Overall, the GPT-optimized AgentPSO skill improves Claude's average accuracy to 58.37\%, outperforming all initial skill baselines.
These results provide preliminary evidence that AgentPSO-learned skills can improve another backbone LLM.
Additional GPT-to-Qwen transfer results are provided in Appendix~\ref{app:gpt-to-qwen}.

\paragraph{Reusing skills with multi-agent inference methods.}
To examine whether AgentPSO-learned skills can serve as reusable skill modules for existing multi-agent frameworks, we plug the optimized skills into MAD and MoA at test time.
Table~\ref{tab:agentpso_test_time_collaboration} shows that replacing the original prompts with AgentPSO-trained skills improves performance on most benchmarks.
For MAD, all agents are instantiated with the global-best skill discovered by AgentPSO, while for MoA, we use different personal-best skills to preserve diversity among proposer agents.
Across both MAD and MoA, replacing the original prompts with AgentPSO-trained skills improves performance on most benchmarks.
This suggests that AgentPSO learns reusable agent skills that are not limited to a single inference procedure, but can be integrated into different multi-agent reasoning frameworks.

\subsection{Analysis}

\paragraph{Effect of velocity on skill stability.}
We further analyze the role of the velocity term ($v$) in AgentPSO.
In the no-velocity variant, we remove the velocity update step in Eq.~(\ref{eq:velocity}) and directly update the skill using Eq.~(\ref{eq:skill}).
Given the skill $s_t$ at iteration $t$ and the updated skill $s_{t+1}$, we compute
$\mathrm{Drift}(s_t, s_{t+1})
=
1 -
\cos\left(
\mathrm{Emb}(s_t),
\mathrm{Emb}(s_{t+1})
\right)$,
where $\mathrm{Emb}(\cdot)$ is computed using OpenAI embedding API\footnote{\url{https://developers.openai.com/api/docs/models/text-embedding-3-small}}.
A larger drift indicates a larger semantic change in the updated skill.
As shown in Table~\ref{tab:velocity_ablation}, removing velocity decreases the final accuracy from 79.50\% to 78.00\%.
Moreover, velocity makes skill evolution smoother, reducing the mean semantic drift and the proportion of high-drift updates, defined as drift $>0.1$.
These results suggest that velocity improves performance and stabilizes evolution by reducing abrupt semantic shifts.

\paragraph{Qualitative analysis of skill evolution.}
We qualitatively examine how AgentPSO updates the velocities and skills of individual agents.
Figure~\ref{fig:velocity_example} and~\ref{fig:skill_evolution_example}
in Appendix~\ref{app:example_of_results} show the evolution of the agent initialized with a Self-Refine skill.
In the early stage, the updates mainly improve answer formatting, such as checking the required output type, using canonical notation, and verifying edge cases.
These changes are primarily driven by the self-reflective direction, which helps the agent correct recurring formatting and validation errors.
In the middle stage, the agent incorporates guidance from the global-best skill, adding higher-level reasoning behaviors such as problem-type classification and method selection.
By the later stage, the skill becomes a hybrid policy that preserves the original Self-Refine structure while integrating global-best abstraction and feedback-derived validation behavior.
As a result, AgentPSO turns simple initial skills into more robust and general reasoning skills.

\begin{figure}[t]
\centering
 
\begin{minipage}[c]{0.42\linewidth}
    \centering
    \includegraphics[width=\linewidth]{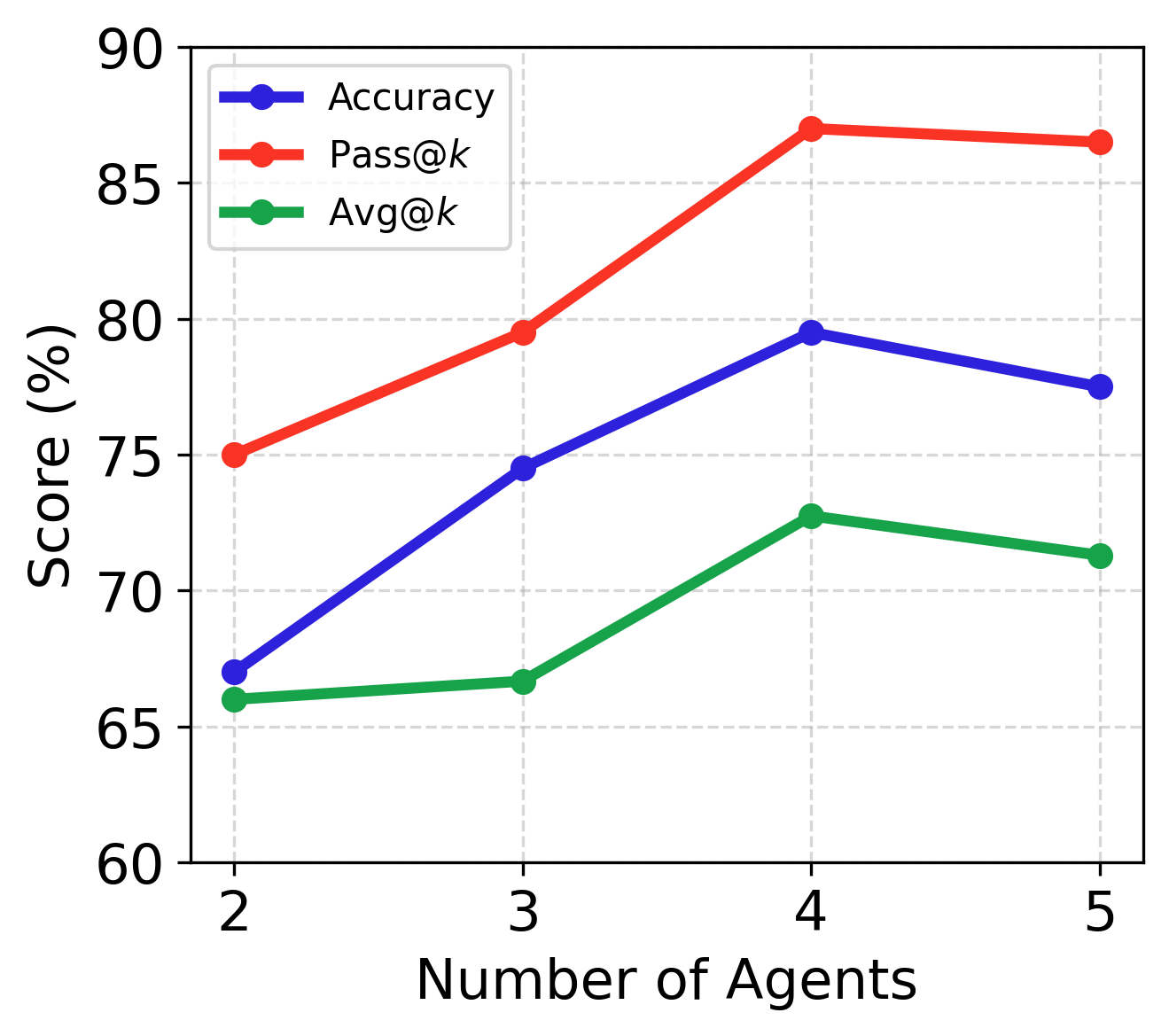}
    \captionof{figure}{\textbf{Effect of the number of agents.}}
    \label{fig:num_agents_all_metrics}
\end{minipage}%
\hfill
\begin{minipage}[c]{0.56\linewidth}
    \centering
    \captionof{table}{\textbf{Ablation study on the update components of AgentPSO on DeepMath.}}
    \label{tab:update_component_ablation}
    \scriptsize
    \setlength{\tabcolsep}{3pt}
    \renewcommand{\arraystretch}{1.15}
    \begin{tabular}{l|cc}
    \toprule
    Input Components & Acc. & Pass@$k$ \\
    \midrule
    $v_i \oplus s_i \oplus d_i \oplus p_i \oplus g$ & 79.50 & 87.00 \\
    $v_i \oplus s_i \oplus p_i \oplus g$ & 72.50 & 80.50 \\
    $v_i \oplus s_i \oplus p_i$ & 73.00 & 80.00 \\
    $v_i \oplus s_i \oplus g$ & 72.00 & 81.50 \\
    $v_i \oplus s_i \oplus d_i$ & 76.00 & 81.50 \\
    \bottomrule
    \end{tabular}
\end{minipage}
 \vspace{-0.2em}
\end{figure}


\subsection{Ablation Study}
\label{sec:ablation_study}

\paragraph{Effect of the number of agents.}
{Figure~\ref{fig:num_agents_all_metrics}}
analyzes how the number of agents affects AgentPSO performance on DeepMath.
Increasing the number of agents from $k=2$ to $k=4$ consistently improves performance.
This shows that agent diversity is beneficial, as a larger population provides more reasoning trajectories and candidate skills for evolution.
However, performance slightly drops when using $k=5$, suggesting that additional agents can introduce redundancy and make the collective update process harder to stabilize.
Therefore, we use $k=4$ as the default number of agents, which provides the best trade-off between diversity and stability.

\paragraph{Contribution of self-reflective direction.}

Table~\ref{tab:update_component_ablation} highlights the importance of the self-reflective direction ($d_i$) in AgentPSO.
When $d_i$ is removed from the full update rule, performance drops substantially, indicating that standard PSO-style guidance alone is insufficient for fully optimizing agent skills.
Personal-best guidance ($p_i$) and global-best guidance ($g$) provide useful signals by preserving and propagating successful skills, but they are most effective when combined with $d_i$.
In contrast, the variant that combines the velocity with the self-reflective direction achieves stronger performance than the variants based only on personal-best or global-best guidance.
This suggests that feedback derived from peer reasoning trajectories provides a more direct and informative signal for improving agent skills.
Additional validation trajectories comparing AgentPSO with and without $d_i$ are provided in Appendix~\ref{app:effect_of_self_reflection_di}.

\section{Conclusion}
\label{sec:conclusion}
We proposed \textbf{AgentPSO}, a PSO-inspired framework that evolves multi-agent reasoning skills in natural-language skill space.
Beyond the standard PSO components of personal-best and global-best guidance, AgentPSO introduces a self-reflective direction that allows agents to extract reusable lessons from peer reasoning trajectories.
By combining these update signals, AgentPSO refines each agent's skill without updating the backbone model parameters.
The learned skills exhibit strong transferability, generalizing across both reasoning benchmarks and backbone LLMs.
This suggests that AgentPSO learns reusable reasoning behaviors instead of task- or model-specific prompting patterns.
Overall, our results show that multi-agent collaboration can serve not only as an inference-time mechanism for answer refinement, but also as a model-parameter-free process for evolving reusable reasoning skills at the population level.
This may improve the reliability and efficiency of reasoning systems by enabling agents to acquire reusable reasoning behaviors without parameter updates.

\paragraph{Limitations.} 
The current experiments focus on reasoning benchmarks. 
Although AgentPSO avoids parameter updates, it requires additional training-time LLM calls to evolve the agent population.
Another limitation is that evolved agent skills may amplify incorrect reasoning patterns if the feedback, peer observations, or validation signals are flawed.
Moreover, AgentPSO optimizes reasoning skills rather than acquiring new domain knowledge.
Future work may combine AgentPSO with stronger verification or retrieval mechanisms, especially for knowledge-intensive domains.

\section*{Acknowledgements}
This work was supported in part by the Institute of Information \& Communications Technology Planning \& Evaluation (IITP) grant funded by the Korea government (MSIT) (RS-2025-02304967, AI Star Fellowship (KAIST)), and in part by a grant from the Korea Health Technology R\&D Project through the Korea Health Industry Development Institute (KHIDI), funded by the Ministry of Health \& Welfare, Republic of Korea (grant number: RS-2025-02213531).

\nocite{}

\bibliography{references}
\bibliographystyle{icml2026}

\newpage
\appendix
\onecolumn

\section{Experimental Details}
\label{app:experiment_details}

\begin{table}[h]
\centering

\begin{minipage}[t]{0.37\linewidth}
\centering
\caption{
\textbf{Hyperparameter settings for AgentPSO.}
}
\label{tab:agentpso_hyperparameters}
\small
\setlength{\tabcolsep}{6pt}
\renewcommand{\arraystretch}{1.15}
\resizebox{\linewidth}{!}{%
\begin{tabular}{l|l}
\toprule
\textbf{Hyperparameter} & \textbf{Value} \\
\midrule
Number of agents & 4 \\
Number of iterations & 10 \\
Training/Validation pool size & 100 / 100 \\
Training/Validation batch size & 10 / 20 \\
Epsilon margin & 0.01 \\
Maximum velocity length & 200 words / 300 tokens \\
Maximum skill length & 1200 words / 800 tokens \\
\bottomrule
\end{tabular}
}
\end{minipage}
\hfill
\begin{minipage}[t]{0.57\linewidth}
\centering
\caption{
\textbf{Dataset statistics used in our experiments.}
}
\label{tab:dataset_statistics}
\small
\setlength{\tabcolsep}{8pt}
\renewcommand{\arraystretch}{1.15}
\resizebox{\linewidth}{!}{%
\begin{tabular}{lccc}
\toprule
\textbf{Dataset} & \textbf{Train} & \textbf{Validation} & \textbf{Test} \\
\midrule
DeepMath~\cite{deepmath} & 100 & 100 & 200 \\
BigBenchHard~\cite{bbh} & 100 & 100 & 200 \\
MATH~\cite{math500} & -- & -- & 500 \\
AIME25~\cite{aime25} & -- & -- & 30 \\
Minerva~\cite{minerva} & -- & -- & 272 \\
\bottomrule
\end{tabular}
}
\end{minipage}

\end{table}

\subsection{Instruction of Agents}
\label{app:skill_modules}
We initialize the agent population with four distinct reasoning skill modules.
Each agent is assigned one initial instruction that encourages a different reasoning behavior.
These initial skills are used to induce diversity among agents before the optimization process begins.
The full instructions are as follows.

\begin{itemize}
    \item \textbf{Agent 1: Chain of Thought.} 
    \textit{Solve the problem step by step.}

    \item \textbf{Agent 2: Step-Back Prompting.} 
    \textit{Before solving, step back and identify the general principle or problem type.}

    \item \textbf{Agent 3: Self-Refine.} 
    \textit{First solve the problem, then review and improve the solution.}

    \item \textbf{Agent 4: Reflection.} 
    \textit{Solve while reflecting on assumptions and possible failure points.}
\end{itemize}

\subsection{Details of Hyperparameters}
\label{app:pso_hyperparameters}

Since AgentPSO operates in natural-language skill space instead of a numeric vector space, we do not sample stochastic PSO coefficients.
Instead, all natural-language update components are incorporated with equal weight.
The primary hyperparameters of AgentPSO control the number of agents, the sizes of the training and validation pools, the batch sizes used during optimization, the epsilon margin for updating best skills, and the maximum lengths of generated velocity and skill descriptions.
Table~\ref{tab:agentpso_hyperparameters} summarizes the hyperparameter settings.

\section{Datasets}
\label{app:datasets}

Table~\ref{tab:dataset_statistics} summarizes the dataset splits used in our experiments.
For DeepMath and BigBenchHard (BBH), we use 100 training samples, 100 validation samples, and 200 test samples.
The training split is used for skill updates, the validation split is used for personal-best and global-best selection, and the test split is used only for final evaluation.
To reduce API cost and computational overhead, we randomly sample 200 test examples, following prior multi-agent reasoning studies~\cite{diverse_mad,multiagent_debate}.
For MATH, AIME25, and Minerva, we perform transfer evaluation without additional training or validation.

\paragraph{Validation scheduler.}
For DeepMath and BBH, the 100 validation samples are divided into five non-overlapping subsets of 20 samples.
At each iteration, validation is conducted on one subset.
If an agent achieves a perfect score on the current validation subset, the next iteration uses a different subset.
This scheduler reduces overfitting to a single saturated validation subset and encourages robust best-skill selection.

\section{Baseline Details}
\label{app:baseline_details}
\paragraph{Vanilla single-agent methods}
\begin{itemize}
    \item Chain-of-Thought Prompting (CoT)~\cite{wei2022chain}: 
    It is instructed to solve the problem step by step, decomposing the problem into intermediate reasoning steps before producing the final answer. 
    We use the same prompt as Agent~1 in Appendix~\ref{app:skill_modules}.
    \item Step-Back Prompting~\cite{stepback}: 
    It is encouraged to first abstract away from the specific problem and identify the underlying principle, concept, or problem type before solving.
    We use the same prompt as Agent~2 in Appendix~\ref{app:skill_modules}.
\end{itemize}

\paragraph{Advanced single-agent methods}
\begin{itemize}
    \item Self-Refine~\cite{selfrefine}: 
    It first generates an initial solution and then reviews and improves its own reasoning before finalizing the answer.
    We use the same prompt as Agent~3 in Appendix~\ref{app:skill_modules}.
    \item Reflection~\cite{shinn2023reflexion}: 
    It solves the problem while reflecting on assumptions, uncertainty, and possible failure points. 
    We use the same prompt as Agent~4 in Appendix~\ref{app:skill_modules}.
    \item Self-Consistency~\cite{selfconsistency}: 
For each problem, we prompt the model with \textit{``Let's think step by step''} and generate $4$ independent reasoning paths using stochastic decoding.
We extract the final answer from each generated path and determine the final prediction by majority voting over the $4$ extracted answers.
    \item Tree-of-Thoughts~\cite{yao2023tot}: 
We implement a Tree-of-Thoughts baseline using breadth-first search.
For each problem, the model expands the current reasoning states by generating $4$ candidate thoughts.
Each partial solution is then evaluated by the same model, which serves as a value function for estimating the promise of the current reasoning state.
We prune the search tree to a beam width of $2$ at each depth and repeat this process up to a maximum depth of $3$.
After the search reaches the maximum depth, the remaining states are completed into final solutions.
The final prediction is selected as the answer from the highest-scoring completed solution.
\end{itemize}

\paragraph{Multi-agent methods}
\begin{itemize}
    \item MAD~\cite{multiagent_debate}: 
We implement a 4-agent, 2-round Multi-Agent Debate baseline.
In the first round, each agent independently solves the problem.
In the second round, each agent observes the other agents' first-round responses and produces a revised final answer.
We use majority voting over the second-round answers as the final prediction.
    \item MoA~\cite{MOA} : 
We implement a 3-layer Mixture-of-Agents baseline.
In Layer 1, $4$ proposer agents independently solve the problem.
In Layer 2, $4$ aggregator agents each receive the original problem and all Layer-1 outputs, and synthesize improved solutions.
In Layer 3, a single final aggregator receives the original problem and all Layer-2 outputs, and produces the final answer.
We use the final aggregator's output as the final prediction.
    \item DMAD~\cite{diverse_mad} : 
We implement a 4-agent, 2-round Diverse Multi-Agent Debate baseline.
Each agent is assigned a distinct reasoning method: Chain-of-Thought, Step-Back Prompting, Self-Refine, or Reflection.
In the first round, each agent independently solves the problem using its assigned method.
In the second round, each agent observes the other agents' first-round solutions and produces a refined answer while maintaining its assigned method.
We use majority voting over the second-round answers as the final prediction.
\end{itemize}

\paragraph{Self-evolving methods}
\begin{itemize}
    \item DSPy~\cite{dspy}: 
We implement a single-agent Chain-of-Thought baseline using DSPy\footnote{\url{https://dspy.ai/}} and optimize its prompt and configuration with MIPROv2.
DSPy represents an LM pipeline as a declarative program with input--output signatures and compiles it by optimizing the prompts and demonstrations used by its modules.
We use MIPROv2 to optimize this single-agent reasoning program on the training set, and use the optimized program directly at test time.
    \item GEPA~\cite{gepa}: 
We implement a single-agent prompt optimization baseline using the GEPA API\footnote{\url{https://gepa-ai.github.io/gepa/}}.
GEPA is a reflective prompt optimizer that improves textual components of an LM system using execution traces, feedback, and LLM-based reflection.
    \item ADAS~\cite{adas}: 
ADAS uses a meta-agent to iteratively generate new executable agent designs from an archive of previously evaluated agents.
We initialize the archive with several hand-designed agents, including Chain-of-Thought, Self-Consistency, Self-Refine, LLM Debate, Quality-Diversity, Step-back Abstraction, and Role Assignment.
Each candidate agent is evaluated on the validation/search set and added to the archive.
After search, we select the best validation agent and evaluate it on the held-out test set.
\end{itemize}

\section{Additional Analysis}

\begin{table*}[t]
\centering
\caption{
\textbf{
Comparison with single-agent, multi-agent, and self-evolving baselines on GPT-5.4-mini.}
All reported values denote accuracy. Each cell reports mean $\pm$ standard deviation over three random seeds.
}
\label{tab:gpt54mini_3seeds_results}
\scriptsize
\setlength{\tabcolsep}{4.0pt}
\renewcommand{\arraystretch}{1.12}
\resizebox{\textwidth}{!}{%
\begin{tabular}{ll|cccccc}
\toprule
\multirow{2}{*}{\textbf{Category}} 
& \multirow{2}{*}{\textbf{Method}}
& \multicolumn{6}{c}{\textbf{GPT-5.4-mini}} \\
\cmidrule(lr){3-8}
& 
& \textbf{DeepMath}
& \textbf{MATH}
& \textbf{AIME25}
& \textbf{Minerva}
& \textbf{BBH}
& \textbf{Avg.} \\
\midrule

\multirow{2}{*}{\shortstack[c]{Vanilla \\ Single-Agent}}
& Chain-of-Thought 
& $68.83 \pm 0.62$ 
& $90.20 \pm 0.16$ 
& $38.89 \pm 1.57$ 
& $38.97 \pm 1.08$ 
& $85.33 \pm 1.03$ 
& $64.44$ \\

& Step-Back Prompting 
& $65.83 \pm 3.40$ 
& $89.87 \pm 0.75$ 
& $47.78 \pm 3.14$ 
& $37.62 \pm 1.65$ 
& $84.17 \pm 1.70$ 
& $65.05$ \\

\midrule

\multirow{4}{*}{\shortstack[c]{Advanced \\ Single-Agent}}
& Self-Refine      
& $67.67 \pm 2.25$ 
& $90.40 \pm 0.86$ 
& $51.11 \pm 1.57$ 
& $36.27 \pm 1.51$ 
& $83.83 \pm 0.94$ 
& $65.86$ \\

& Reflection       
& $68.67 \pm 2.72$ 
& $90.60 \pm 0.91$ 
& $44.44 \pm 4.16$ 
& $38.73 \pm 2.13$ 
& $83.17 \pm 0.62$ 
& $65.12$ \\

& Self-Consistency 
& $68.33 \pm 1.04$ 
& $89.13 \pm 0.42$ 
& $54.44 \pm 5.09$ 
& $28.19 \pm 1.12$ 
& $86.33 \pm 0.76$ 
& $65.28$ \\

& ToT       
& $63.17 \pm 1.76$ 
& $81.60 \pm 0.80$ 
& $52.22 \pm 3.85$ 
& $30.15 \pm 2.30$ 
& $78.83 \pm 1.04$ 
& $61.19$ \\

\midrule

\multirow{3}{*}{\shortstack[c]{Multi-Agent}}
& MAD       
& $70.00 \pm 1.00$ 
& $\textbf{95.93} \pm 0.12$ 
& $\textbf{56.67} \pm 0.00$ 
& $33.82 \pm 1.10$ 
& $82.33 \pm 1.44$ 
& $67.75$ \\

& MoA       
& $71.00 \pm 1.00$ 
& $95.53 \pm 0.12$ 
& $55.56 \pm 1.92$ 
& $28.68 \pm 1.91$ 
& $78.67 \pm 2.47$ 
& $65.89$ \\

& DMAD     
& $71.50 \pm 1.32$ 
& $95.67 \pm 0.61$ 
& $55.56 \pm 1.92$ 
& $35.17 \pm 0.56$ 
& $83.67 \pm 1.89$ 
& $68.31$ \\

\midrule

\multirow{3}{*}{\shortstack[c]{Self-Evolving}}
& DSPy     
& $74.67 \pm 3.21$ 
& $92.27 \pm 0.42$ 
& $52.22 \pm 5.09$ 
& $36.52 \pm 0.93$ 
& $83.50 \pm 5.07$ 
& $67.84$ \\

& GEPA    
& $\textbf{77.83} \pm 1.76$ 
& $91.67 \pm 1.03$ 
& $48.89 \pm 5.09$ 
& $39.95 \pm 3.08$ 
& $82.17 \pm 4.04$ 
& $68.10$ \\

& ADAS     
& $62.33 \pm 2.31$
& $79.67 \pm 2.01$ 
& $54.44 \pm 1.92$ 
& $30.76 \pm 2.22$ 
& $84.17 \pm 1.04$ 
& $62.27$ \\

\midrule

\multirow{1}{*}{\shortstack[c]{AgentPSO}}
& AgentPSO 
& $77.00 \pm 3.50$ 
& $94.27 \pm 0.58$ 
& $53.33 \pm 3.33$ 
& $\textbf{41.67} \pm 2.02$ 
& $\textbf{86.67} \pm 1.76$ 
& $\textbf{70.59}$ \\

\bottomrule
\end{tabular}
}
\vspace{-2em}
\end{table*}

\begin{figure*}[t]
\centering
\begin{minipage}[t]{0.6\linewidth}
    \centering
    \vspace{0pt}
    \includegraphics[width=0.95\linewidth]{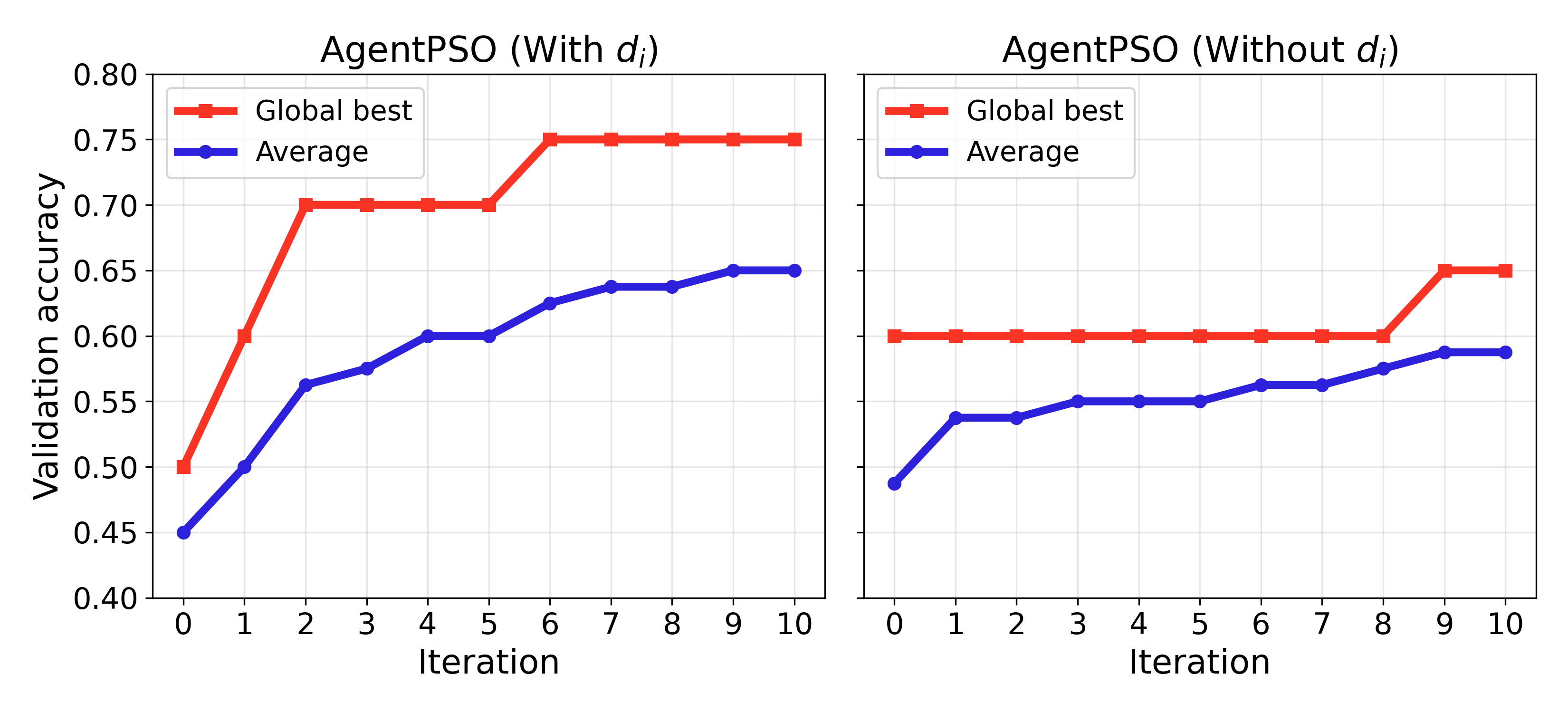}
    \caption{\textbf{Validation accuracy of the global-best and average accuracy over iterations.}
    The global-best trajectory improves steadily, while the variant without $d_i$ shows only limited progress.}
    \label{fig:test_with_without_di}
\end{minipage}
\hfill
\begin{minipage}[t]{0.38\linewidth}
    \centering
    \vspace{0pt}
    \includegraphics[width=0.85\linewidth]{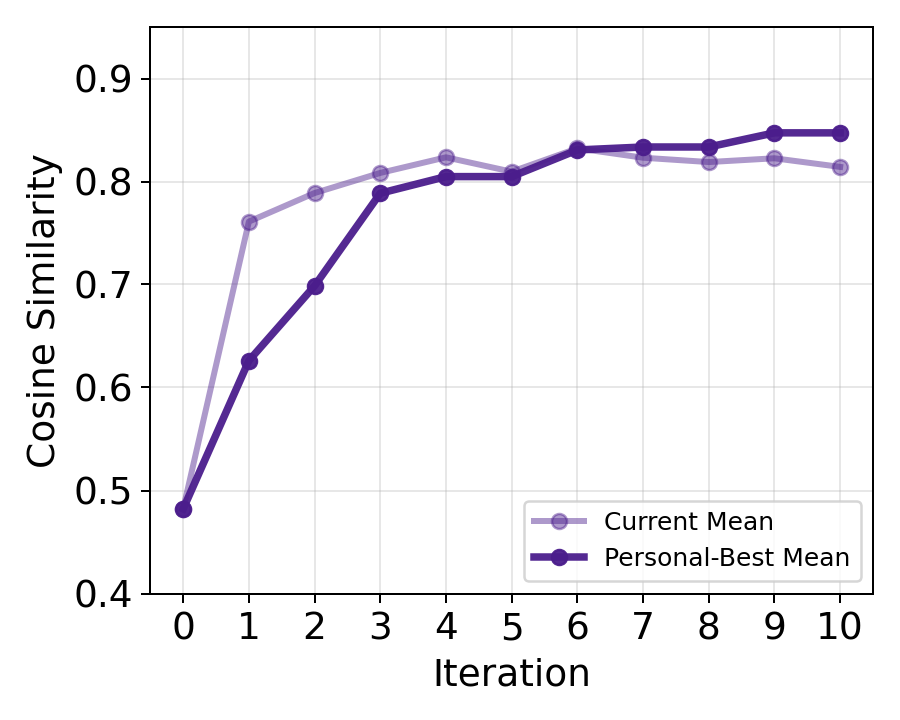}
    \caption{
    \textbf{Semantic alignment with the final global-best skill.}
Agent skills become increasingly similar to the final global-best skill over iterations.
    }
    \label{fig:similarity_agents_graph}
\end{minipage}
\vspace{0.2em}
\end{figure*}

\subsection{
Seed-Level Robustness Analysis
}
\label{app:repeat-seed-results}
To assess whether the improvements of AgentPSO are robust across different random seeds, we repeat the GPT-5.4-mini experiments with three random seeds. Table~\ref{tab:gpt54mini_3seeds_results} reports the mean accuracy and standard deviation for each benchmark. 
The results are consistent with the GPT-5.4-mini results in Table~\ref{tab:combined_main_qwen_results}, and AgentPSO achieves the best average accuracy among single-agent, multi-agent, and self-evolving baselines.

\subsection{
Effect of the self-reflective direction
}
\label{app:effect_of_self_reflection_di}


To isolate the contribution of the self-reflective direction $d_i$, Figure~\ref{fig:test_with_without_di} compares the full AgentPSO update (With $d_i$) and a PSO-style variant (Without $d_i$).
While both variants benefit from PSO-style guidance, AgentPSO with $d_i$ achieves better performance than the variant without $d_i$ throughout optimization.
In contrast, removing $d_i$ leads to weaker progress, suggesting that personal-best and global-best guidance alone are not sufficient to fully drive skill evolution.
These results indicate that the self-reflective direction $d_i$ helps each agent learn from its own mistakes and from peer agents' reasoning, leading to better skill updates over iterations.

\subsection{Skill alignment dynamics} \label{app:similarity_agents}
To examine how the agent population evolves relative to the final best skill
in the DeepMath experiment, 
we measure the mean cosine similarity between each agent's skill and the final global-best skill across iterations.
Embeddings are computed using the OpenAI embedding API\footnote{\url{https://developers.openai.com/api/docs/models/text-embedding-3-small}} and averaged over agents using DeepMath benchmark.
As shown in Figure~\ref{fig:similarity_agents_graph}, the similarity rises in the early iterations and then plateaus well below 1, indicating that agents partially converge toward the global-best skill but do not collapse onto it.
This behavior is consistent with the intended role of AgentPSO: the population-level guidance steers agents toward effective regions of the skill space, while the self-reflective direction and per-agent personal-best preserve diversity across the swarm.

\subsection{
Inference latency
}
\label{app:inference_latency}

\begin{table}[t]
\centering
\caption{
\textbf{Inference latency comparison.}
Training time is reported in minutes and measured over the entire training set.
Latency is measured on 30 DeepMath test examples.
Relative latency is normalized by the single-agent baseline.
}
\label{tab:latency_comparison}
\small
\setlength{\tabcolsep}{7pt}
\renewcommand{\arraystretch}{1.15}
\begin{tabular}{l|c|cccc}
\toprule
\textbf{Method} 
& \textbf{Train Time (m)} 
& \textbf{Calls/case}
& \textbf{Avg. Inference time/case (s)}
& \textbf{Relative latency} \\
\midrule
Single Agent & - & 1 & 5.40 & 1.00$\times$ \\
MAD & - & 8 & 42.72 &  7.92$\times$ \\
MoA & - & 9 & 52.45 & 9.72$\times$ \\
\midrule
DSPy & 26.4  & 1  &  5.35 &  0.99$\times$ \\
GEPA & 19.17  & 1  & 4.52  &  0.84$\times$ \\
ADAS & 1469.1  & 4  & 18.26 & 3.38$\times$ \\
\midrule
AgentPSO & 190.55 & 4 & 22.95 &  4.25$\times$ \\
\bottomrule
\end{tabular}
\vspace{-0.2mm}
\end{table}



We compare the training and inference costs of AgentPSO with single-agent, multi-agent, and self-evolving baselines.
Table~\ref{tab:latency_comparison} reports the training time on the full training set and the inference latency on DeepMath test examples.
The average inference time per case denotes the mean time across the 30 examples required to solve one problem.
Relative latency is normalized by the single-agent baseline, where $1.00\times$ corresponds to the latency of a single-agent inference.

For training-free methods, the difference in latency mainly comes from the number of LLM calls required at test time.
The single-agent baseline uses one LLM call per case, while MAD and MoA require eight and nine calls per case, respectively.
In contrast, AgentPSO requires only four calls per case at test time because the optimized agents solve the problem independently and their answers are aggregated by majority voting.
Although AgentPSO requires a separate training stage, it avoids costly test-time debate or layer-wise aggregation.
As a result, AgentPSO has an average inference time of 22.95 seconds per case, substantially lower than MAD and MoA, which require 42.72 and 52.45 seconds per case.

We also compare AgentPSO with self-evolving methods that require an optimization stage over the training set.
DSPy and GEPA have lower inference latency because they use a single optimized program or prompt at test time, but their final average performance is lower than AgentPSO.
ADAS, on the other hand, requires a much larger training cost while still using multiple LLM calls at inference time.
This shows that AgentPSO can retain the benefits of multi-agent reasoning while reducing test-time computational overhead.


\begin{table}[t]
\centering
\caption{\textbf{Cross-model transfer of AgentPSO-learned skills from GPT to Qwen3-32B.}}
\label{tab:gpt_skill_to_qwen}
\small
\setlength{\tabcolsep}{7pt}
\renewcommand{\arraystretch}{1.15}
\begin{tabular}{l|ccccc|c}
\toprule
\textbf{Method}
& \textbf{DeepMath}
& \textbf{MATH}
& \textbf{AIME25}
& \textbf{Minerva}
& \textbf{BBH}
& \textbf{Avg.} \\
\midrule
Chain-of-Thought     & 60.00 & 77.20 & 10.00 & 37.50 & 71.50 & 51.24 \\
Step-Back Prompting  & 51.50 & 63.80 & 6.67  & 37.87 & 66.50 & 45.27 \\
Self-Refine          & 62.00 & 78.40 & 6.67  & \textbf{41.54} & 71.00 & 51.92 \\
Reflection           & 57.00 & 74.00 & 13.33 & 41.18 & 74.00 & 51.90 \\
\midrule
\textbf{AgentPSO (GPT$\rightarrow$Qwen)}
& \textbf{80.00} & \textbf{86.60} & \textbf{20.00} & 37.87 & \textbf{75.00} & \textbf{59.89} \\
\bottomrule
\end{tabular}
\end{table}

\subsection{Cross-model transfer: GPT to Qwen3-32B} \label{app:gpt-to-qwen}

To test whether GPT-discovered skills can transfer to an open-weight backbone, we directly apply the best skill optimized with GPT to Qwen3-32B without additional training or adaptation.
Table~\ref{tab:gpt_skill_to_qwen} compares this transferred skill with Qwen3-32B single-agent baselines using the initial skill prompts.
The single-agent baselines use Qwen3-32B with the initial skill prompts, while AgentPSO~(GPT$\rightarrow$Qwen) uses the GPT-optimized AgentPSO skill.
The transferred skill improves average accuracy to 59.89\%, outperforming all initial skill baselines.
These results suggest that AgentPSO-learned skills capture transferable reasoning behaviors that can improve another backbone LLM without additional optimization.

\subsection{Additional medical benchmark}
\label{app:medical_benchmark_results}

\begin{table}[t]
\centering
\caption{
\textbf{AgentPSO on medical question-answering benchmarks.}
AgentPSO is optimized on MedXpertQA and evaluated on MedXpertQA and MedBullets.
Results show strong pass@$k$ performance, indicating improved coverage of correct solutions among agents.
}
\label{tab:medical_result}
\small
\setlength{\tabcolsep}{7pt}
\renewcommand{\arraystretch}{1.15}
\begin{tabular}{l|cc|cc}
\toprule
\multirow{2}{*}{\textbf{Method}} 
& \multicolumn{2}{c|}{\textbf{MedXpert}}
& \multicolumn{2}{c}{\textbf{MedBullets}} \\
& \textbf{Accuracy} 
& \textbf{Pass@$k$} 
& \textbf{Accuracy} 
& \textbf{Pass@$k$} \\
\midrule
Chain-of-Thought
& 20.50 & - & 82.50 & - \\
Step-Back Prompting
& 20.00 & - & 82.50 & - \\
Self-Refine
& 22.00 & - & 84.00 & - \\
Reflection
& 20.50 & - & 84.00 & - \\
MAD
& 43.50 & 47.50 & 86.00 & 87.50  \\
\midrule
AgentPSO
& 36.00 & 52.50 & 84.50 & 89.00 \\
\bottomrule
\end{tabular}
\end{table}
To further examine whether AgentPSO can be applied beyond mathematical and general reasoning tasks, we conduct additional experiments on medical question-answering benchmarks.
We use MedXpertQA~\cite{zuo2025medxpertqa} and MedBullets~\cite{chen2024benchmarkinglargelanguagemodels}, which contain challenging medical exam-style questions requiring domain-specific reasoning.
AgentPSO is optimized on MedXpertQA using the same four-agent setting as in the main experiments, with 100 training examples and 200 test examples.
We then evaluate the optimized agents on both MedXpertQA and MedBullets.

Table~\ref{tab:medical_result} summarizes the results.
On MedXpertQA and MedBullets, AgentPSO improves pass@$k$ over MAD, indicating that the optimized agent population is more likely to contain at least one correct solution among its agents.
However, its majority-voting accuracy is lower than MAD.

Unlike mathematical or general reasoning benchmarks,
medical QA is often constrained more by the model's domain-specific knowledge reliability than by reasoning diversity alone.
Since AgentPSO updates reasoning skills instead of updating underlying medical knowledge, the evolved agents may explore more diverse solutions while still failing to form a reliable majority when the required clinical knowledge is not reliably represented.
Thus, these results show that AgentPSO can improve answer coverage, as reflected by higher pass@$k$, but skill-only optimization is less effective for improving majority-vote accuracy in knowledge-intensive domains.

\section{Example of Results}
\subsection{Example of Velocity}
\label{app:example_of_velocity}
This section provides a qualitative example of how the velocity evolves during AgentPSO optimization.
We use Agent 3 as a representative example, which is initialized with the Self-Refine skill.

Figure~\ref{fig:velocity_example} shows the velocity descriptions produced at different training iterations.
Unlike the skill itself, the velocity represents the \emph{update direction} that guides how the agent should revise its current reasoning behavior at each step.
Each instruction in the velocity is annotated according to its main source of influence:
\Dtag{} indicates content introduced by the feedback-guided self-reflective direction,
\GBtag{} indicates influence from the global-best skill,
\PBtag{} indicates influence from the agent's personal-best skill,
\Stag{} indicates influence from the agent's current skill, and
\Vtag{} indicates content carried over from the previous velocity.
When the personal-best and current skills are identical, we use \PGBtag{} because their contributions cannot be separated.
These annotations illustrate how AgentPSO constructs each update direction by combining self-reflection feedback, guidance from the global-best and personal-best skill, the agent's own skill state, and accumulated momentum from previous velocities.

Overall, this example shows that the velocity in AgentPSO is not a simple copy of either the global-best or personal-best skill.
Instead, it functions as a compositional update signal that synthesizes multiple information sources into a coherent revision direction.
This behavior highlights the role of velocity as the mechanism that translates feedback and peer influence into gradual but meaningful agent evolution.

\begin{table}[t]
\centering
\caption{
\textbf{Qualitative trajectory of agent skill.}
Across iterations, the skill evolves from generic self-refinement and output-format control toward structured problem-type classification, edge-case verification, proof-to-answer calibration, and strict benchmark-format compliance.
}
\label{tab:skill_refinement_compact}
\small
\setlength{\tabcolsep}{5pt}
\renewcommand{\arraystretch}{1.15}
\begin{tabular}{p{0.12\linewidth}|p{0.78\linewidth}}
\toprule
\textbf{Iteration} & \textbf{Newly Emphasized Behavior} \\
\midrule
Iter. 1 
& Self-review, final-format validation, removal of extraneous output, and direct final-answer generation. \\
Iter. 3
& Early problem-type classification, method selection, and explicit handling of symbolic, algebraic, geometric, and boundary-sensitive cases. \\
Iter. 9 
& Internal self-refinement, proof-to-answer calibration, domain/existence/uniqueness checks, and stricter benchmark-format compliance. \\
\bottomrule
\end{tabular}
\end{table}

\subsection{Example of Skill Evolution}
\label{app:example_of_results}
This section provides a qualitative example of how an agent skill evolves during AgentPSO optimization.
We again use Agent 3 as an example, which is initialized with the Self-Refine skill.

Figure~\ref{fig:skill_evolution_example} shows the detailed skill descriptions saved at different iterations.
Each instruction is annotated according to its main source of influence:
\PBtag{} indicates instructions inherited from the previous personal-best skill,
\GBtag{} indicates instructions influenced by the global-best skill, and
\Dtag{} indicates instructions introduced by the feedback-guided self-reflective direction.
These annotations help show how AgentPSO combines personal-best preservation, global-best guidance, and reflective feedback when updating an agent skill.

Table~\ref{tab:skill_refinement_compact} provides a compact summary of the same trajectory.
Across iterations, the skill evolves from a generic self-refinement instruction into a more structured reasoning policy.
In the early stage, the agent mainly learns to align its response with the required answer format, such as checking the expected output type and removing extraneous content.
In the middle stage, it begins to incorporate task-level reasoning strategies, including problem-type classification and method selection.
By the later stage, the skill further emphasizes domain and edge-case verification, proof-to-answer calibration, and strict benchmark-format compliance.
Overall, this example shows that AgentPSO does not simply replace an agent's initial prompt, but progressively refines it by integrating useful behaviors from multiple update sources.

\begin{figure}[p]
\centering
\begin{minipage}{1\linewidth}
\begin{skillbox}{Velocity at Iteration 1}
    \PGBtag{} \Dtag{} Revise the workflow to solve first, then run a stricter final validation pass.
    \PGBtag{} \Dtag{} Keep the current review habit, but make the last check more deliberate about benchmark alignment and answer exactness.
    \Dtag{} Confirm whether the task expects an expression, equation, or plain value.
    \Dtag{} Use the canonical symbol/notation.
    \Dtag{} Remove any stray wording, braces, JSON, or metadata before responding.
    \Dtag{} Preserve the strong mathematical reasoning.
    \Dtag{} When the answer is simple, prefer a direct, problem\-specific argument over a looser derivation.
    \Dtag{} In symbolic or boundary\-sensitive problems, especially algebra/cardinality and geometry, explicitly verify the final form and edge cases.
    \Dtag{} \GBtag{} Ensure that the response matches the required format exactly.
\end{skillbox}

\vspace{0.1em}
\begin{skillbox}{Velocity at Iteration 3}
    \PGBtag{} \Vtag{} \Dtag{} Revise as Self-Refine with a stricter final validation pass.
    \PGBtag{} Solve first, then re-check the answer.
    \Dtag{} \PGBtag{} \Vtag{} Check the answer against the exact required output type and benchmark notation.
    \PGBtag{} Keep the self-correcting review habit.
    \PGBtag{} \Dtag{} Verify whether the response must be a plain value, expression, or equation.
    \Dtag{} \PGBtag{} \Vtag{} Strip all commentary, JSON, braces, metadata, and unsupported equalities unless explicitly requested.
    \Dtag{} For choice or parameter problems, return only the required symbol or scalar.
    \PGBtag{} \Vtag{} For symbolic, algebraic, geometric, or boundary-sensitive tasks, preserve careful reasoning internally but output only the canonical final form.
    \PGBtag{} \Vtag{} \Dtag{} Check edge cases, existence/domain conditions, and notation consistency.
    \PGBtag{} Start by classifying the problem type and choosing the most fitting method.
    \Dtag{} \PGBtag{} \Vtag{} Perform a brief final check for correctness, minimality, and exact token match before responding.
\end{skillbox}

\vspace{0.1em}

\begin{skillbox}{Velocity at Iteration 9}
    \Stag{} \GBtag{} Revise as Self-Refine with a stronger step-back start.
    \GBtag{} \Stag{} \Vtag{} First classify the task, identify the governing principle, and determine the exact required output form before solving.
    \GBtag{} \Stag{} Use the standard method for the problem class.
    \Stag{} \GBtag{} \Vtag{} Keep all reasoning internal.
    \Stag{} \PBtag{} Solve directly, then run a strict self-check.
    \Stag{} \PBtag{} \Vtag{} Verify correctness, edge cases, domain/existence/uniqueness, and boundary conditions when relevant.
    \Dtag{} Tighten proof-to-answer calibration.
    \Dtag{} Pay special attention to abstract algebra/field-theory and quotient/classification problems.
    \Dtag{} Verify claims against the exact prompt and avoid overgeneralizing from partial structure.
    \GBtag{} \Stag{} \PBtag{} \Vtag{} Strengthen final-form control so the response matches the benchmark's expected shape exactly.
    \PBtag{} \GBtag{} \Vtag{} Distinguish whether the final answer should be a value, equation, or canonical expression.
    \PBtag{} \Stag{} \GBtag{} \Vtag{} Use minimal notation, with no extra commentary, proof, metadata, or unsupported equalities.
    \Dtag{} Add a final consistency pass for sign, counting, and maximal/subquotient-style arguments.
    \Dtag{} \Stag{} \Vtag{} Ensure symbolic or yes/no prompts return only the requested form.
\end{skillbox}

\end{minipage}
\caption{
\textbf{Example of velocity across AgentPSO training iterations.} 
The velocity is evolved from an empty string.
\PBtag{} denotes influence from Agent's personal-best skill, 
\GBtag{} denotes influence from the global-best skill, 
\Dtag{} denotes the self-reflective direction at the corresponding iteration, 
\Stag{} denotes influence from Agent's current skill, and 
\Vtag{} denotes accumulated influence from the previous velocity.
When the personal-best and current skills are identical, we use \PGBtag{} because their contributions cannot be separated.
}
\label{fig:velocity_example}
\end{figure}

\begin{figure}[p]
\centering
\begin{minipage}{1\linewidth}
\begin{skillbox}{Skill at Iteration 1}
\begin{itemize}[leftmargin=1.2em]
    \item \PBtag{} Solve the problem first, then review and refine the solution.
    \item \PBtag{} Keep the reasoning strong and self-correcting.
    \item \GBtag{} \Dtag{} In the final pass, validate strict alignment with the task's required output format.
    \item \Dtag{} Check whether the answer should be an expression, equation, or plain value, and use the canonical notation/symbols.
    \item \Dtag{} Remove any stray wording, braces, JSON, metadata, or other extraneous content before responding.
    \item \PBtag{} When the solution is simple, give a direct, problem-specific answer rather than an unnecessarily long derivation.
    \item \Dtag{} For symbolic, boundary-sensitive, or geometry/algebra problems, explicitly verify edge cases and the final form.
    \item \GBtag{} \Dtag{} Return only the final answer in the exact form the task expects.
\end{itemize}
\end{skillbox}

\vspace{0.1em}

\begin{skillbox}{Skill at Iteration 3}
\begin{itemize}[leftmargin=1.2em]
    \item \PBtag{} Solve the problem first, then self-review and refine the result.
    \item \PBtag{} Preserve a disciplined, self-correcting approach aligned with the task.
    \item \GBtag{} Classify the problem type early and choose the most appropriate method.
    \item \PBtag{} \Dtag{} In the final pass, verify the exact required output type: plain value, expression, or equation.
    \item \GBtag{} \Dtag{} Use canonical notation and the benchmark's expected symbols; ensure domain, existence, and edge-case conditions are satisfied.
    \item \PBtag{} For simple tasks, answer directly without extra derivation.
    \item \Dtag{} For symbolic, algebraic, geometric, or boundary-sensitive tasks, check edge cases and confirm the final form is correct.
    \item \PBtag{} \Dtag{} Remove commentary, metadata, braces, JSON, and unsupported equalities unless explicitly requested.
    \item \Dtag{} For choice or parameter problems, return only the required symbol or scalar.
    \item \PBtag{} \GBtag{} Output only the final answer in the exact form requested.
\end{itemize}
\end{skillbox}

\vspace{0.1em}

\begin{skillbox}{Skill at Iteration 9}
\begin{itemize}[leftmargin=1.2em, itemsep=1pt, topsep=2pt]
    \item \PBtag{} Self-Refine: solve internally, then self-check and refine before responding.
    \item \GBtag{} First classify the task, identify the governing principle, and determine the exact required output form.
    \item \GBtag{} Use the standard method for the problem class and reason carefully, but keep all reasoning internal.
    \item \PBtag{} \Dtag{} Solve directly, then verify domain, existence, uniqueness, edge cases, boundary conditions, and any relevant sign/counting/maximality issues.
    \item \Dtag{} Tighten proof-to-answer calibration: make sure the result matches the exact prompt and do not overgeneralize from partial structure, especially in abstract algebra, field theory, and quotient/classification problems.
    \item \PBtag{} Perform a strict final validation for correctness, completeness, and format compliance.
    \item \GBtag{} Prefer canonical notation and the simplest correct final expression.
    \item \GBtag{} Match the benchmark's expected shape exactly---value, equation, or canonical form---with minimal notation.
    \item \GBtag{} Output only the final answer, with no extra commentary, scaffolding, proof, metadata, or unsupported equalities.
    \item \Dtag{} For yes/no or symbolic prompts, return only the requested form.
\end{itemize}
\end{skillbox}

\end{minipage}
\caption{
\textbf{Example of skill evolution across AgentPSO training iterations.} The skill is evolved from ``\textit{First solve the problem, then review and improve the solution.}''.
\PBtag{} denotes instructions inherited from the previous personal-best skill, 
\GBtag{} denotes instructions influenced by the global-best skill, and 
\Dtag{} denotes instructions introduced by the feedback-guided reflective direction.
}
\label{fig:skill_evolution_example}
\end{figure}

\section{Prompts in AgentPSO}
\label{app:prompts}
Figures~\ref{fig:prompt_problem_solve}--\ref{fig:prompt_update_skill} summarize the prompts used in AgentPSO.
The pipeline first prompts each agent to solve the problem using its current skill, then uses the agents' outputs to derive a self-reflective update direction.
This direction is combined with the previous velocity, personal-best skill, and global-best skill to generate a PSO-guided natural-language velocity.
The generated velocity is then applied to the current skill to produce the next skill while preserving the agent's identity.

\begin{figure*}[t]
    \centering
    \begin{tcolorbox}[colback=gray!5!white, colframe=gray!75!black, title={
    Problem Solving Prompt
    }, fonttitle=\bfseries]
    \small
  You are a mathematics competition agent.

  Base rules:\\
  - Solve the provided MATH problem.\\
  - The answer may be a number, expression, tuple, interval, or simplified LaTeX expression.\\
  - Do not invent missing problem details.\\
  - Follow the current skill file below.\\

  Current skill file:\\
  \{skill\_text\}\\

  Solve this problem using only your current skill.\\
  
  Problem:\\
  \{question\}\\

  Return exactly one JSON object:\\
  \{\\
      ``agent\_id'': \{agent\_id\},\\
      ``reasoning'': ``...'',\\
      ``answer'': ``...''\\
  \}
    \end{tcolorbox}
\caption{\textbf{Problem-solving prompt used by each agent.}
Given the current skill file, each agent independently solves the input problem and returns its reasoning trace and final answer in JSON format.}
    \label{fig:prompt_problem_solve}
\end{figure*}

\begin{figure*}[t]
    \centering
    \begin{tcolorbox}[colback=gray!5!white, colframe=gray!75!black, title={
    Generate Self-reflective Direction~Eq.\eqref{eq:reflect}
    }, fonttitle=\bfseries]
    \small
  Current agent skill:\\
  \{current\_skill\}\\

  Agent's own reasoning traces and answers:\\
  \{own\_outputs\_json\}\\

  Other agents' reasoning traces and answers, including correctness:\\
  \{peer\_outputs\_json\}\\

  Instruction:\\
  Analyze the agent's performance compared with peers.\\
  Identify general reasoning improvements.\\
  Do not overfit to a single problem.\\
  Do not rewrite the skill yet.\\
  Return only the update direction.
    \end{tcolorbox}
\caption{\textbf{Self-reflective direction generation prompt.}
This prompt generates an update direction by comparing the current agent's performance with peer agents, aiming to identify general reasoning improvements.}
    \label{fig:prompt_d_i}
\end{figure*}

\begin{figure*}[t]
    \centering
    \begin{tcolorbox}[colback=gray!5!white, colframe=gray!75!black, title={
    PSO-Guided Velocity~Eq.\eqref{eq:velocity}
    }, fonttitle=\bfseries]
    \small
  Agent identity to preserve:\\
  \{agent\_identity\}\\

  Previous velocity v\_i:\\
  \{previous\_velocity\}\\

  Self-reflective direction d\_i:\\
  \{direction\}\\

  Current skill s\_i:\\
  \{current\_skill\}\\

  Personal best skill p\_i:\\
  \{personal\_best\_skill\}\\

  Global best skill g:\\
  \{global\_best\_skill\}\\

  Instruction:\\
  Combine the previous velocity, self-reflective direction, lessons from the personal best, and lessons from the global best.\\
  Focus on generalizable improvements.\\
  Do not copy the personal best or global best directly.\\
  Preserve the agent's identity.\\
  Return a concise natural-language velocity of at most \{max\_velocity\_words\} words.
    \end{tcolorbox}
\caption{\textbf{Prompt for generating the PSO-guided velocity $v_i^{t+1}$.}
Given the previous velocity, the self-reflective direction, the personal-best skill, and the global-best skill, the model generates a concise natural-language velocity that captures generalizable improvements while preserving the agent's identity.}
    \label{fig:prompt_update_velocity}
\end{figure*}

\begin{figure*}[t]
    \centering
    \begin{tcolorbox}[colback=gray!5!white, colframe=gray!75!black, title={
     Skill Update from Velocity~Eq.\eqref{eq:skill}
    }, fonttitle=\bfseries]
    \small
  Agent identity to preserve:\\
  \{agent\_identity\}\\

  Current skill s\_i:\\
  \{current\_skill\}\\

  Velocity v\_i:\\
  \{velocity\}\\

  Instruction:\\
  Rewrite the skill according to the velocity.\\
  Keep the skill concise and general.\\
  Preserve the agent's original role.\\
  Remove redundant, overly specific, or contradictory instructions.\\
  Use at most 10 bullet points and at most \{max\_skill\_words\} words.\\
  Do not copy another skill verbatim.\\
  Return only the updated skill.\\
    \end{tcolorbox}
\caption{\textbf{Prompt for applying the velocity to the skill.}
Given the current skill and the PSO-guided velocity, the model rewrites the skill into an updated form that preserves the agent's identity while incorporating concise and generalizable improvements.}
    \label{fig:prompt_update_skill}
\end{figure*}


\end{document}